\documentclass[letterpaper]{article} 
\usepackage{aaai24}  
\usepackage{times}  
\usepackage{helvet}  
\usepackage{courier}  
\usepackage[hyphens]{url}  
\usepackage{graphicx} 
\usepackage{natbib}  
\usepackage{caption} 
\usepackage{algorithm}
\usepackage{algorithmic}

\usepackage{newfloat}
\usepackage{listings}
\DeclareCaptionStyle{ruled}{labelfont=normalfont,labelsep=colon,strut=off} 
\floatstyle{ruled}
\newfloat{listing}{tb}{lst}{}
\floatname{listing}{Listing}
\usepackage{tcolorbox}
\usepackage{refcount}

\title{Acceleron: A Tool to Accelerate Research Ideation}
\author{
    Harshit Nigam\textsuperscript{\rm 1},
    Manasi Patwardhan\textsuperscript{\rm 2},
    Lovekesh Vig\textsuperscript{\rm 3},
    Gautam Shroff\textsuperscript{\rm 4}
}
\affiliations{
    TCS Research \\
    \textsuperscript{\rm 1}h.nigam@tcs.com, \textsuperscript{\rm 2}manasi.patwardhan@tcs.com, \textsuperscript{\rm 3}lovekesh.vig@tcs.com, \textsuperscript{\rm 4}gautam.shroff@tcs.com
}

\usepackage{bibentry}

\begin{document}

\maketitle

\begin{abstract}
Several tools have recently been proposed for assisting researchers during various stages of the research life-cycle. However, these primarily concentrate on tasks such as retrieving and recommending relevant literature, reviewing and critiquing the draft, and writing of research manuscripts. Our investigation reveals a significant gap in availability of tools specifically designed to assist researchers during the challenging ideation phase of the research life-cycle. To aid with research ideation, we propose `Acceleron', a research accelerator for different phases of the research life cycle, and which is specially designed to aid the ideation process. Acceleron  guides researchers through the formulation of a comprehensive research proposal, encompassing a novel research problem. The proposals motivation is validated for novelty by identifying gaps in the existing literature and suggesting a plausible list of techniques to solve the proposed problem. We leverage the reasoning and domain-specific skills of Large Language Models (LLMs) to create an agent-based architecture incorporating colleague and mentor personas for LLMs. The LLM agents emulate the ideation process undertaken by researchers, engaging researchers in an interactive fashion to aid in the development of the research proposal.  Notably, our tool addresses challenges inherent in LLMs, such as hallucinations, implements a two-stage aspect-based retrieval to manage precision-recall trade-offs, and tackles issues of unanswerability. To showcase the ideation capabilities of `Acceleron', we illustrate the execution of our motivation validation and method synthesis workflows on proposals from the machine learning and natural language processing domain, given as an input by 3 distinct researchers. Our observations and evaluations provided by the researchers illustrate the efficacy of the tool in terms of assisting researchers with appropriate inputs at distinct stages and thus leading to improved time efficiency.
\end{abstract}

\section{Introduction}
With fast-paced research happening in every field, we are witnessing an exponential growth in the number of scientific articles and research papers on the web. It is difficult for an individual researcher or a small research team to keep abreast of the relevant advances amidst this information explosion. This has a downstream impact on the ability to be consistently appraised and ensure novelty of a proposed solution at various stages of the research life cycle. Thus there is an urgent need for a tools that can aid researchers to 1) understand, evaluate and incorporate the latest developments in the literature and 2) Formulate/Modify the current proposed solution accordingly to ensure novelty and impact.

\begin{figure}[t]
\begin{center}

\includegraphics[width=1.05\columnwidth,height=0.83\columnwidth]{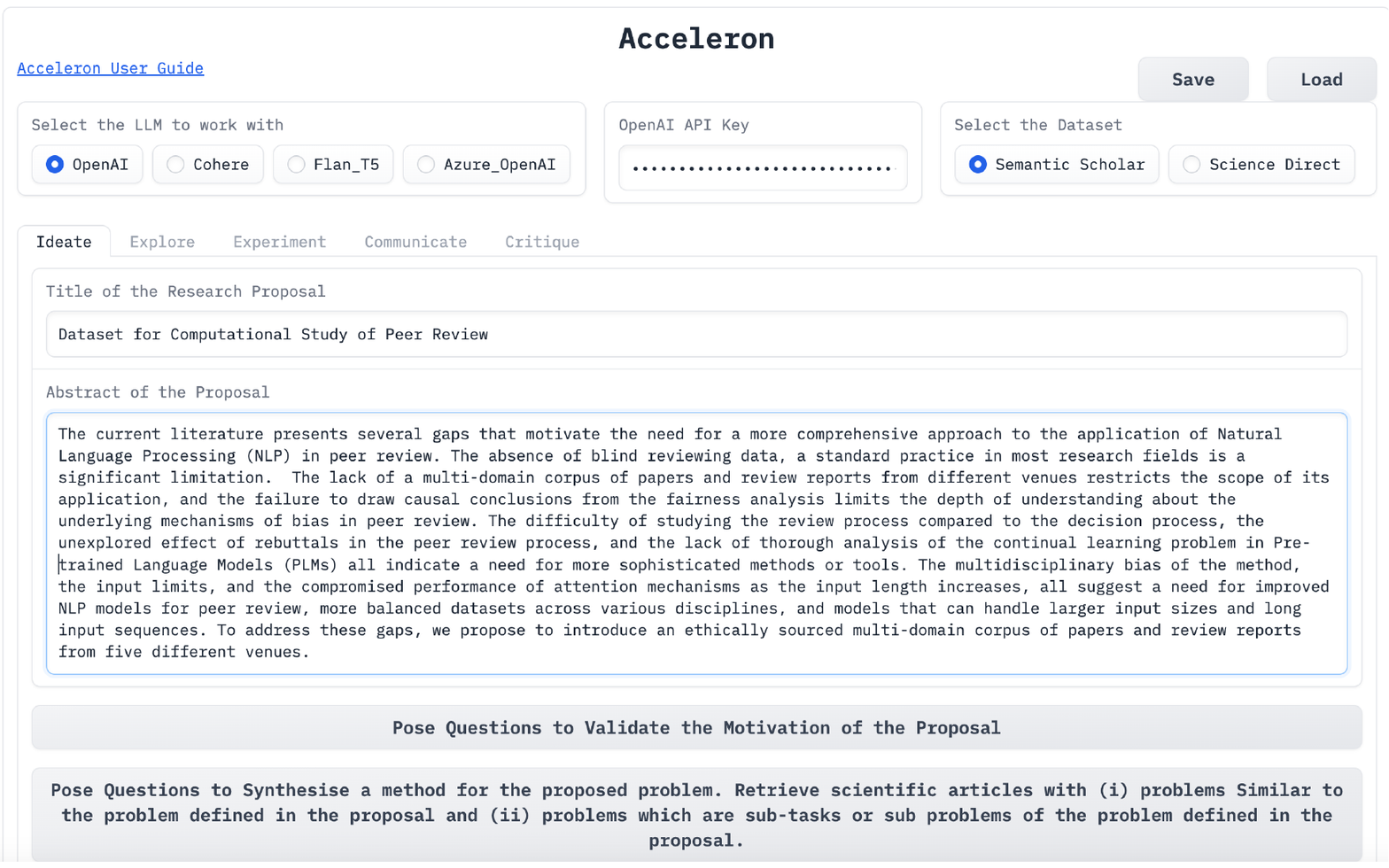}     \caption{Acceleron Interface}
\label{fig:acc}
\end{center} 
\end{figure}

Existing tools facilitate searching of research papers relevant to one's topic of interest based on a query (e.g. Elicit\footnote{\label{elicit}https://elicit.com}, Raxter.io\footnote{\label{raxter}https://raxter.io/}, SciSpace\footnote{\label{SciSpace}https://typeset.io}), keywords  (e.g. Litmaps\footnote{\label{Litmaps}https://www.litmaps.com}, Iris.ai\footnote{\label{iris}https://iris.ai}), paper titles (e.g. inciteful\footnote{\label{inciteful}https://inciteful.xyz} ), through citation graphs (e.g. Litmaps\footnotemark[\getrefnumber{Litmaps}], inciteful\footnotemark[\getrefnumber{inciteful}], Research Rabbit\footnote{\label{rr}https://researchrabbitapp.com}, Connected Papers\footnote{https://www.connectedpapers.com}) or facts and insights (e.g. Scite\footnote{\label{scite}https://scite.ai/}) as inputs. They further help in (i) extracting relevant information from the papers with predefined templates (e.g. Elicit\footnotemark[\getrefnumber{elicit}], Iris.ai\footnotemark[\getrefnumber{iris}]) or based on user queries (e.g. SciSpace\footnotemark[\getrefnumber{SciSpace}]), (ii) sharing the papers in collaborative fashion (e.g. Litmaps\footnotemark[\getrefnumber{Litmaps}], Raxter.io\footnotemark[\getrefnumber{raxter}], Research Rabbit\footnotemark[\getrefnumber{rr}]), (iii) notifying researchers about articles relevant to their searches, tracking trends (e.g. Scite\footnotemark[\getrefnumber{scite}], Research Rabbit\footnotemark[\getrefnumber{rr}]), etc. Some tools  act as the facilitator for reading or writing (e.g. Paperpal\footnote{https://paperpal.com}) manuscripts by allowing highlighting, extracting and documenting important aspects of the paper (e.g. Raxter.io\footnotemark[\getrefnumber{raxter}]), communicating in more interactive fashion to analyze the paper in depth (e.g. SciSpace\footnotemark[\getrefnumber{SciSpace}], Iris.ai\footnotemark[\getrefnumber{iris}]), recommending relevant quality citations (e.g. Scite\footnotemark[\getrefnumber{scite}], Raxter.io\footnotemark[\getrefnumber{raxter}]) or summarizing papers (e.g. Elicit\footnotemark[\getrefnumber{elicit}], Iris.ai\footnotemark[\getrefnumber{iris}]). Thus, most of the existing tools focus on notifying and recommending researchers with relevant literature, facilitate exploration of existing literature and/or writing research manuscripts. Researchers have also proposed learning representations for retrieval of relevant scientific articles \cite{Singh2022SciRepEvalAM,Cohan2020SPECTERDR,stendorff2022NeighborhoodCL,Mysore2021MultiVectorMW},  literature Review Generation \cite{Hu2014AutomaticGO,Kasanishi2023SciReviewGenAL,Chen2021CapturingRB}, Question Answering over scientific articles \cite{Saikh2022ScienceQAAN,qasper:2021,Lee2023QASAAQ}, Scientific document summarization \cite{Hayashi2020WhatsNS}, citation recommendation \cite{Ali2021GlobalCR,Ali2022SPRSMNSP,Medic2023ParagraphlevelCR} citation intent detection \cite{Cohan2019StructuralSF,Berrebbi2022GraphCiteCI,Roman2021CitationIC,Lauscher2021MultiCiteMR}, critical review and rebuttal generation \cite{Ruggeri2022ArgSciChatAD,d2023aries,Kennard2021DISAPEREAD,Dycke2022NLPeerAU,Wu2022IncorporatingPR}, etc. However, to the best of our knowledge, no tool or no approach in the literature facilitates a researcher during the most arduous ideation stage of the research life-cycle. Ideation involves: (i) Analyzing the existing literature to critically evaluate the motivation behind the research problem a researcher is trying to address to ensure that the mentioned research gap(s) still exist(s), (ii) Reformulating the proposed research problem and objectives based on the validation stage output and re-identification of research gaps, (iii) Identifying analogous research problems or sub-problems addressed in the literature and utilizing their solutions, available in the literature, to derive a set-of approaches or synthesizing a set-of plausible methods as a solution to the problem, (iv) Designing experimentation strategy for the given problem and selected methodology. 

Most of the tasks involved in research require domain expertise and complex reasoning skills. The recent advancement in Large Language Models (LLMs) and Generative Artificial Intelligence (GenAI) has made it possible to partially automate some of these tasks \cite{Liu2023ReviewerGPTAE,Liang2023CanLL,Zhang2023WhenLL,Lahiri2023CitePromptUP,Kunnath2023PromptingSF}. In this work, we propose `Acceleron’ (Figure \ref{fig:acc}), a tool to accelerate the research life cycle. By exploiting the reasoning and domain specific skills of LLM based agents, the goal of the tool is to assist with research activities, alleviating the burden of researchers. The aim is not to replace a researcher, but to assist the researcher by providing relevant inputs in an interactive fashion, at various stages of research life cycle, viz. ideate, explore, codify, experiment, communicate, critique, inform; to expedite the process of meeting the research objectives. In this paper we are focusing on the `ideate' module of the tool, which facilitates the ideation of a research problem, specified by the researcher. We expect the researcher to provide a short paragraph along with the title of the research problem the researcher plan to address, along with the description of what motivates the researcher to solve the problem. With LLM powered \textit{mentor} and \textit{colleague} agents, the tool, in an interactive fashion, helps the researcher to develop the research proposal consisting of a validated motivation, a well-defined research problem focusing of research gaps in the literature, a proposed approach selected from a set-of plausible synthesized methods and possible set-of experiments to be conducted to evaluate the approach for the research problem. To the best of our knowledge, we are the first ones to mimic the ideation process, followed by researchers, using the LLM agents. We demonstrate the ideation power of `Acceleron' by illustrating examples from natural language processing. 



\section{System Architecture}

\begin{figure}[t]
\begin{center}

\includegraphics[width=\columnwidth]{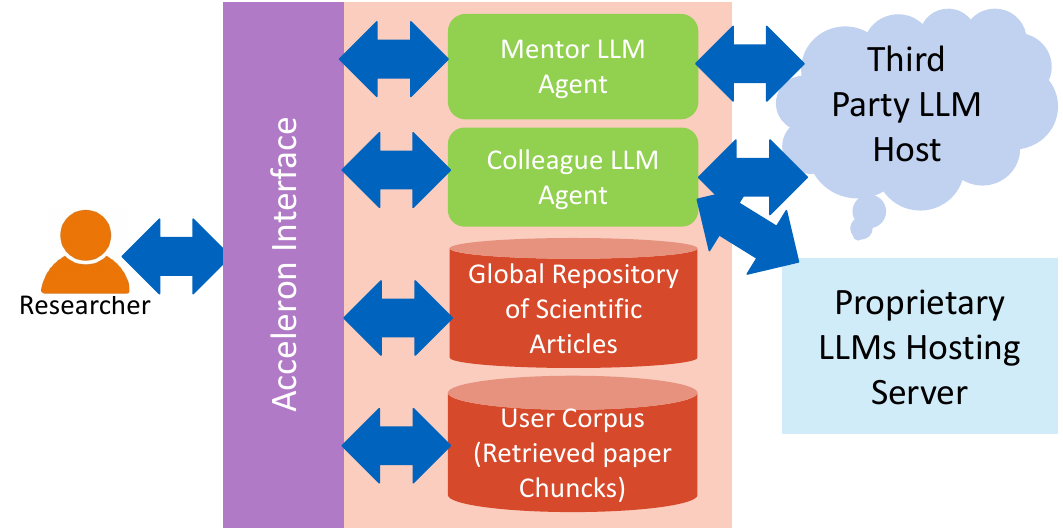}     \caption{System Architecture}
\label{fig:pipeline}
\end{center} 
\end{figure}
Acceleron provides a web-based interface for researchers to interact. The system architecture is illustrated in Figure \ref{fig:pipeline}. We define an LLM Agent based architecture \cite{Wang2023ASO}, with agents of two distinct types of profiles or personas\footnote{We use Langchain framework https://www.langchain.com for implementation}. A \textit{Colleague} persona\footnote{OpenAPI's GPT-turbo-3.5 model} performs less complex tasks including extraction of relevant information from user inputs, generation of relevant questions from extracted information or retrieval of relevant data from scientific documents. Whereas, \textit{mentor} persona\footnote{OpenAPI's GPT4 model} performs more complex tasks requiring reasoning such as understanding the limitations or gaps of the existing work, identifying problems similar to the problem discussed in the proposal, identifying sub-tasks of the problem being solved in the proposal, solving similar problems and/or sub-tasks to synthesize a solution to the proposed problem and re-write the proposal given a plausible set-of approaches or possible limitations of related work. The architecture is flexible such that the LLM agents can interact with (i) LLMs using API calls\footnote{LLMs designed by OpenAI https://platform.openai.com/docs/api-reference or Cohere https://cohere.com} or (ii) open-source LLMs which reside on an internal hosting server. 

We expect to have a global repository which is a vector store\footnote{We use pinecone https://www.pinecone.io/} of domain specific scientific articles\footnote{We use more than 2 million scientific articles in semantic scholar fetched using S2ORC dataset \cite{lo-etal-2020-s2orc} as the global repository} which are indexed by the Specter embeddings \cite{Cohan2020SPECTERDR} produced using the paper's title and abstract. We also have a User Specific corpus which has chunks of all the retrieved papers relevant to the current proposal the researcher is working on. The paper chunks are created with our in-house parser\footnote{We have built the parser using pdfminer https://pypi.org/project/pdfminer/} treating paragraphs as semantic segments. If a paragraph does not fit into the the maximum token length of LLM agents, while chunking it is further split to fit into the maximum token length. The chunks are further converted to vector embeddings and indexed\footnote{We use FAISS indexing https://github.com/facebookresearch/faiss} for efficient retrieval based on semantic similarity with a query. This user corpus acts as a shared `memory' for the LLM agents.

\section{Approach}
In this section we describe the ideation process adopted by Acceleron which simulates the ideation process followed by researchers using LLM agents. The process involves interaction between a researcher and the LLM agents, where the LLM agents perform actions based on the feedback received by the researcher or another agent. The process takes a proposal as an input from a researcher with a research problem description specified at a high level along with the motivation behind the problem. The output of the ideation process is the updated proposal with a (i) Validated motivation or updated research problem by identifying gaps mentioned in the literature addressing the motivation (ii) Plausible methods to address the research problem.  The overall ideation task is split into two workflows: (i) Motivation Validation and (ii) Method Synthesis. The detailed prompts for the steps in each of the workflow are illustrated in the Appendix Section \ref{sec:prompts}.
\subsection{Motivation Validation Workflow}
\begin{figure}[t]
\begin{center}

\includegraphics[width=\columnwidth]{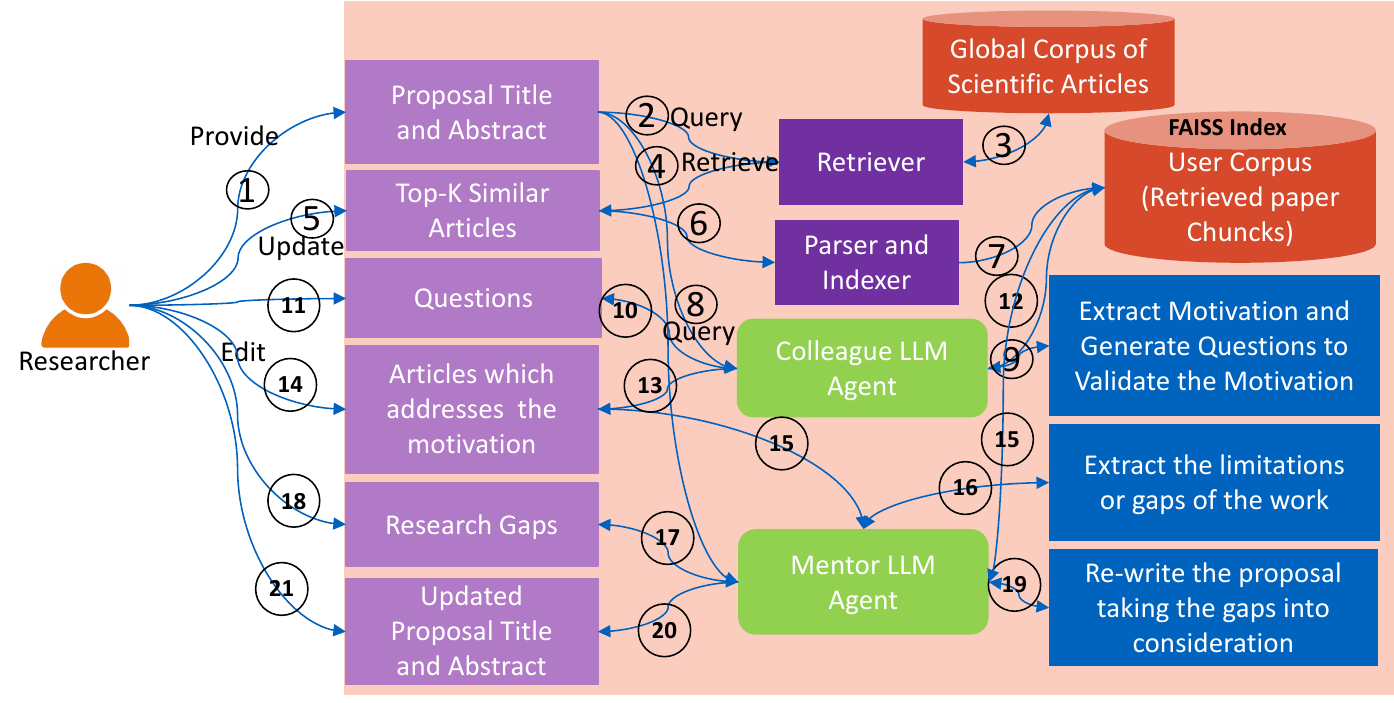}     \caption{Motivation Validation Workflow}
\label{fig:valid}
\end{center} 
\end{figure}

This is the first phase of the ideation process. The workflow is elaborated in Figure \ref{fig:valid}. The steps of the workflow are explained in detail here: 1. The researcher provides the title and abstract of the proposal elaborating the motivation behind the proposal and a high level description of the problem statement the research wishes to solve. 2. The retriever functionality of Acceleron uses this title and abstract of the proposal as a query and gets a vector representation\footnote{Specter \cite{Cohan2020SPECTERDR} } of the same 3. This representation is used to retrieve top-K articles similar to the contents of the proposal from the global corpus of scientific articles.  4. The top-K articles along with a description of the relevance of each article to the proposal is shown to the researcher. 5. These articles are editable by the researcher who can delete articles they find irrelevant or by manually adding relevant articles. 6. The parser and indexer functionality chunks the finalized set-of papers in paragraphs (semantic segments) 7. The chunks are stored in the user corpus with appropriate indexing mechanisms  8. The \textit{colleague} agent fetches the proposal title and abstract provided by the researcher 9. It extracts the motivation out of the proposal (Prompt 1) and generates a list of questions to be posed on the shortlisted scientific articles to validate the motivation of the proposal (Prompt 2). The questions are binary and formulated such that if, for a scientific article, the answer to the question is `yes', then the article is already addressing the motivation of the proposal mentioned by that question. For example, if the researcher proposes to develop a technique to solve a novel aspect of a problem,  a question would be generated of the form `Does the research paper address that specific aspect of the problem?'. If this question is answered as `yes' by a scientific article then it implies that the article addresses that aspect of the problem and hence the motivation behind the study is weak or invalidated. 10. The generated set-of questions are shown to the researcher 11. The researcher can edit the questions by updating the format of the questions, deleting questions that the researcher deems irrelevant or adding missing relevant questions 12. For each question and a retrieved paper stored in the user corpus, the \textit{colleague} agent retrieves the chunks of the paper relevant to that question and tries to answer that question using Retrieval Augmented Generation (RAG) (Prompt 3). The answer can be `yes', `no' or unanswerable along with an explanation. 13.  If all the papers answer either `no' or `unanswerable' for all the questions generated to validate the motivation; it indicates that the existing literature is not addressing the motivation behind the proposal and hence this phase is ended with a comment shown to the researcher that the motivation of the proposal is validated. Otherwise, the question-research paper pairs with only `yes' as an answer, along with explanation are shown to the researcher. 14. The researcher can edit this output in terms of removing papers which he doesn't agree to address the question based on the explanation provided. If the answers are hallucinated, this step enables the researcher to remove such papers. 15. The \textit{mentor} agent uses the chunks of each of the shortlisted paper, the original proposal and the description of the prior question addressing the motivation 16. It extracts the limitations or gaps of each of the papers(Prompt 4), which has been identifying to be addressing the motivation of the proposal, such that the gaps can help redefining the problem in the proposal. 17. The research gaps are shown to the researcher 18. The researcher can ignore the gaps  found to be irrelevant and selects some of the gaps that address part of the research problem. If the researcher does not agree with any of the specified research gaps, they can add their own set of research gaps 19. The \textit{mentor} agent uses these gaps along with the original proposal to re-formulate the motivation(Prompt 5) and the problem statement of the proposal to address the new research gaps. 20. The modified proposal is presented to the researcher(Prompt 6).  21. The modified proposal can be edited by the researcher to finalize the same or he can reject the edits and go back to the prior proposal. This workflow can be applied to the proposal in an iterative manner. Meaning,  the motivation behind the resultant updated proposal can be again validated by initiating the same workflow. This can be executed in an iterative fashion, until there are no scientific articles retrieved to be addressing the motivation behind the proposal, validating the novelty of the proposal. 

\subsection{Method Synthesis Workflow}

\begin{figure}[t]
\begin{center}

\includegraphics[width=\columnwidth]{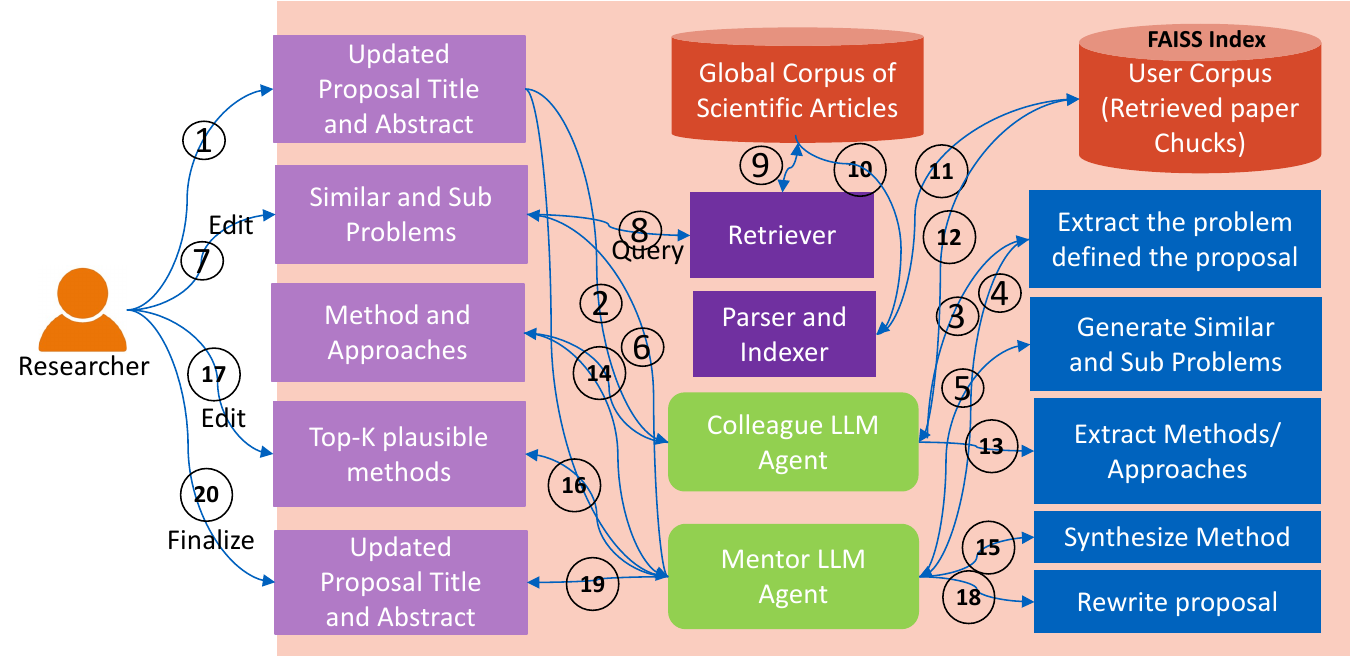}     \caption{Method Synthesis Workflow}
\label{fig:synth}
\end{center} 
\end{figure}

The steps of the method synthesis workflow are illustrated in Figure \ref{fig:synth} and are elaborated here: 1. The method synthesis phase begins with the proposal whose motivation is validated based on reviewing the literature and accepted by the researcher 2. The \textit{colleague} agent takes the proposal as an input 3. It extracts the problem defined in the proposal(Prompt 7) 4. The \textit{mentor} agent takes this problem as an input and 5. uses its` parametric knowledge to generate a plausible set-of similar research problems(Prompt 8). For example, if the problem defined in the proposal is `To design a reference-free evaluation metric for question answering task'. A similar problem can be `To come up with an evaluation metric for text summarization which can have multiple possible reference summaries'. The \textit{mentor} agent also uses its parametric knowledge to decompose the problem defined in the proposal into sub-tasks(if any) (Prompt 11). For example, the problem of `question answering over scientific papers' can be decomposed into `extraction of text from PDF document of the scientific paper', `segmenting the paper and storing it for efficient retrieval',  `retrieval of  paragraphs from the paper related to the question', `answering the question using the retrieved paragraphs as context', `evaluating the retrieved paragraphs' and `evaluating the answers'.  6. The generated similar and sub problems are shown to the researcher 7. The researcher can edit them by removing the ones found to be irrelevant, updating them  or adding missing ones. This helps in removing the problems hallucinated by the agent 8. Each of the edited similar problem or sub-problem is used as a query to retrieve scientific articles from the global corpus which address those respective problems. 10. The retrieved scientific articles are parsed and chunked and 11. are stored in the user corpus. 12. For each retrieved paper, the \textit{colleague} agent extracts the paragraphs which talks about the method or approach taken by the paper to solve the respective problem and 13. generates a consolidate list of similar or sub problem and solution pairs.(Prompt 9, 10, 12) 14. This list is showcased to the researcher who makes edits to the same 15. The extracted problem of the proposal along with the consolidated list of similar or sub problems along with their solutions in the literature is provided to the \textit{mentor} agent.(Prompt 14, 15) 15.  The \textit{mentor} agent uses this information along with its' parametric knowledge to synthesize a list of plausible methods to solve the problem defined in the proposal.(Prompt 16) 16. The list of plausible methods is shown to the researcher.(Prompt 13) 17. Researcher can choose a subset of these methods which they deem most plausible and further edit these if need be. 18. The updated list is provided to the \textit{mentor} agent along with the original proposal. The \textit{mentor} agent re-writes the proposal including these methods.(Prompt 17) 19. This updated proposal is shown to the researcher 20. The researcher can further edit the proposal and finalize the same.

\section{Novel Components}

\textbf{LLM agents for research ideation:}
To the best of our knowledge, ours is the first LLM agent based tool which assists in the complex task of ideation for research. We have devised with two novel portfolios for LLMS, viz., \textit{colleague} and \textit{mentor}, allocating less complex tasks to the \textit{colleague} agent and more complex reasoning based tasks to the \textit{mentor} agent. The user corpus acts as the shared memory for the agents, whereas the agents perform fixed set of actions at various stages of the workflow based on the provided inputs as discussed in the prior sections. Rather than using a costly LLM like GPT4 for all the tasks involved in the workflows; dividing the tasks as per the difficulty level and leveraging less costly LLM such as GPT-turbo-3.5 for colleague agent, performing less complex tasks, provides a cost-effective solution for workflows.

\textbf{Mitigation of hallucination:}
Hallucination is one of the major difficulties of using LLMs for knowledge based tasks \cite{Zhang2023SirensSI,Wang2023SurveyOF}. We mitigate this problem using a two-fold solution: (i) There are retrieval augmented components of the workflows, viz. the motivation validation workflow poses questions generated to validate the motivation of the proposal on the retrieved articles stored in the user corpus  or extract limitations of the articles which address the proposal motivation or the method synthesis workflow extracts approaches used to solve similar or sub problems from the retrieved articles. For these retrieval augmented tasks through proper prompt engineering, we ensure that the answers are provided by restricting the knowledge to the retrieved context only. We observe this helps to mitigate hallucinations. (ii) There are components of the workflows which rely on parametric knowledge of LLMs, for example the motivation validation involves re-writing the proposal and the method synthesis involves generating similar sub problems for the problem defined in the proposal and synthesizing methods. For these tasks the output can not be restricted to the provided input. In such cases, there is a higher chance of hallucinated outputs. For such scenarios, we ensure mitigation of hallucinated outputs, by keeping the system semi-automated and allowing user-interactions at every step to edit or delete hallucinated outputs. Moreover at every stage of the workflow, the LLM agents are asked to justify their outputs and the provided justification is exposed to the researcher through the interface. This forces the model to apply Chain-of-Thoughts (COT) \cite{Wei2022ChainOT}  and allows the researcher to validate the output and check if it is in sync with the justification provided. This assists in alleviating the effect of hallucinations.

\textbf{Two-stage aspect based retrieval:} 
The global corpus contains a large number of scientific articles stored with the Specter embedding of the title and abstract of the papers. The title and abstract of the papers contains information about motivation and problem statement of the papers and a high level mention of the methodology and the results. For ideation we  require more in-depth information from the papers across various aspects such as methodology, limitations, etc. To achieve this we perform retrieval in two stages. In motivation validation workflow, we first retrieve top-K papers from the global corpus with the proposal as the query and high value of K for good recall. This allows us to have a set-of papers with similar motivation and problem statement to that of the proposal. These papers are chunked and stored in the user corpus for further aspect based retrieval, such as papers with similar motivation to that of the proposal and paper paragraphs mentioning the research gaps of these papers. In method synthesis workflow, we first retrieve top-K papers from the global corpus with similar sub problem statements as the query and high value of K for good recall. This allows us to have a set-of papers with problems similar to the problem described in the proposal or similar to any of the sub-tasks of the problem described in the proposal. These papers are chunked and stored in the user corpus for further aspect based retrieval such as extracting the approaches of the papers. Note that keeping high-recall for the first stage of retrieval ensures coverage of papers, whereas for the second stage we favor more precise outcomes for aspect based retrieval.

\textbf{Introduction of Unanswerability:}
The output of aspect based retrieval is always top-K paragraphs from the retrieved and chunked papers. We keep the value of K low to get more precise retrieval for the given aspect based query. However, there is a possibility that the retrieved paragraphs do not have the answer to the query (the query is unanswerable). For example, in the motivation validation workflow the retrieved paragraphs from the papers do not answer the question of whether the paper addresses a specific motivation of the proposal and does not specify the limitations of the paper which would help to refine the problem defined in the proposal. Similarly, for the method synthesis workflow the retrieved paragraphs may not have an approach to solve a similar problem. In such cases, the LLM based agents check the relevancy of retrieved paragraphs for the given query and identifies the query as `unanswerable' in case if all the retrieved paragraphs are irrelevant, avoiding irrelevant outputs. Allowing unasnwerability also assists in reduction of hallucinations.

\section{Qualitative Analysis of the Workflows}
In the absence of an appropriate dataset for the tasks relevant to the ideation process, we provide a qualitative analysis of the workflows with 3 proposals from distinct researchers, specifically in the domain of Artificial Intelligence (AI), Machine Learning (ML) and Natural Language Processing (NLP). The topics of these proposals are: (i) Datasets for Computational Study of Peer Reviews  (ii) Topic-based citation retrieval for research proposal and (iii) Reference-free evaluation metric for retrieval augmented question answering  
These researchers are from our lab, working on distinct research problems, for which they intend to write proposals. We use Semantic Scholar data fetched using S2ORC dataset \cite{lo-etal-2020-s2orc} as our global repository, which has a variety of papers in the AI, ML and NLP domain. We utilize the logging functionality of `Acceleron' to keep track of the interactions between the researcher and the LLM Agents and derive the following observations.

The  abstract of the proposal with topic `\textit{Dataset for Computational Study of Peer Reviews}' is: `\textit{Peer review constitutes a core component of scholarly publishing and demands substantial expertise and training, and is susceptible to errors and biases. Various applications of NLP for peer reviewing assistance aim to support reviewers in this complex process, but the lack of clearly licensed datasets and multi-domain corpora prevent the systematic study of NLP for peer review. To remedy this, we plan to introduce an ethically sourced multi-domain corpus of papers and review reports from five different venues.}'. With the proposal topic and abstract as an input, the motivation validation workflow generates the question: `Is the research paper addressing the lack of clearly licensed datasets for studying natural language processing for peer review?'. As a part of this workflow, to validate the motivation of the proposal, the Colleague LLM agent poses this question on the chunks of top-50 scientific articles, retrieved to be similar to the proposal topic and abstract and stored in the user corpus. Out of these 50 scientific articles, 5 articles are retrieved with `yes' as a answer, out of which the researcher agrees to the justification provided for four and disagrees with the justification of one. Following are the 4 research articles with a valid justification: (i) NLPeer: A Unified Resource for the Computational Study of Peer Review (ii) A Dataset of Peer Reviews (PeerRead): Collection, Insights and NLP Applications (iii) Investigating Fairness Disparities in Peer Review: A Language Model Enhanced Approach (iv) MOPRD: A multidisciplinary open peer review dataset. Whereas, the article with title `What Can We Do to Improve Peer Review in NLP?' agrees with the motivation of the proposal, however, does not address the same. As the next part of the motivation validation workflow, the colleague LLM agent extracts the following research gaps from the scientific articles: (i) NLPeer: A Unified Resource for the Computational Study of Peer Review:  (a) The paper does not contain blind reviewing data which is a standard practice in most research fields, (b) The paper does not perform extensive hyper parameter search and tuning of the models, which could limit the effectiveness of the models, (c) The paper acknowledges the risk of "lazy reading" where reviewers might only read the paragraphs suggested by the model, but does not provide a solution to prevent this. (ii) A Dataset of Peer Reviews (PeerRead): Collection, Insights and NLP Applications: (a) The models used in the research are relatively simple, which may limit their effectiveness in complex peer review scenarios, (b) The paper leave room for further research in areas such as demographic biases in accept/reject decisions, (c) The research does not provide a multidomain corpus of papers and review reports from different venues, limiting the scope of its application. (iii) Investigating Fairness Disparities in Peer Review, A Language Model Enhanced Approach: (a) The research paper does not draw any causal conclusions from the fairness analysis, limiting the depth of understanding about the underlying mechanisms of bias in peer review, (b) The research paper does not investigate the effect of rebuttals in the peer review process, which could be a significant factor in the final decision on a paper, (c) The research paper does not provide a thorough analysis of the continual learning problem in Pretrained Language Models (PLMs), which could be crucial for improving automatic review generation. (iv) MOPRD: A multidisciplinary open peer review dataset: (a) The research paper acknowledges that the multidisciplinary bias of their method could be a concern, indicating a need for more balanced datasets across various disciplines, (b) The paper points out that the input limits still exist as the maximum length of the input text only reaches 16,384 tokens, indicating a need for models that can handle larger input sizes, (c) The paper mentions that the performance of attention mechanisms will be compromised as the input length increases, indicating a need for more efficient attention mechanisms or alternative models for handling long input sequences. This showcases the quality of outputs provided by the workflow in terms of identifying the research gaps in the existing articles, partially eliminating the need of detailed literature survey of these articles to find gaps and thus alleviating research efforts.

We further observe that the researcher selects a subset of these research gaps, she finds to be relevant. The mentor LLM agent further uses these selected gaps to re-write the proposal. Following is the updated proposal provided as an outcome of the workflow taking into account the selected research gaps. \textit{The current literature presents several gaps that motivate the need for a more comprehensive approach to the application of Natural Language Processing (NLP) in peer review. The absence of blind reviewing data, a standard practice in most research fields is a significant limitation.  The lack of a multidomain corpus of papers and review reports from different venues restricts the scope of its application. The difficulty of studying the review process compared to the decision process, the unexplored effect of rebuttals in the peer review process indicate a need for more sophisticated methods or tools. The multidisciplinary bias of the method, the input limits of Language Models, and the compromised performance of attention mechanisms as the input length increases, all suggest a need for improved NLP models for peer review, more balanced datasets across various disciplines, and models that can handle larger input sizes and long input sequences. To address these gaps, we propose to introduce an ethically sourced multidomain corpus of papers and review reports from five different venues.}  The mentor LLM agent takes into consideration the following selected research gaps and introduces them in the revised version of the proposal: (i) availability of multi-disciplinary, multi-venue, blind-review corpus, (ii) no studies of effect of rebuttals (iii) context limits of LMs to tackle long reviews of existing work in this domain. We further observe that the researcher makes a very few edits to the the revised proposal abstract and accepts the same. The total time required for validating the motivation and updating the proposal abstract accordingly is substantially ($\sim$5x for this proposal as mentioned by the researcher) reduced by this workflow.

We receive an input from another researcher with a proposal titled `\textit{Topic-based citation retrieval for research proposal}' and the corresponding abstract `\textit{Retrieval of research articles pertinent to a given query represents a thoroughly investigated research challenge. Typically, queries take the form of a title and abstract of a research article, or a specific sentence or paragraph from an existing research article requiring citation. However, existing approaches presuppose the availability of a well-constructed manuscript, an assumption that is inappropriate during the initial research proposal writing stage. At this initial phase, researchers seek pertinent literature for citing in their proposals, often focusing on specific topics or intents and further build the proposal. In this work, we aim to tackle the issue of topic-based citation retrieval for research proposals. We anticipate researchers providing the title and abstract of their research proposals, encompassing elements such as the research gap, problem statement, and a high-level overview of the proposed methodology and experiments. Additionally, researchers will furnish a list of topics for which relevant scientific articles need to be retrieved. Our proposed algorithm intends not only to fetch research articles pertinent to the given proposal from a corpus, but also to establish a crucial many-to-many mapping between these articles and the specified topics.}' The colleague LLM agent generates the following questions for validation of the motivation: 1. Is the research paper specifically addressing the retrieval of research articles relevant to a topic of a research proposal? and 2. Is the research paper developing a technique to map research articles to specified topics in research proposals?. Out of top-50 research articles used to validate the motivation of the proposal by posing the above mentioned questions, the following four got retrieved to be answering as `yes' to at the least one of the above questions and thus invalidating the motivation behind the proposal: 1 Citation Recommendation: Approaches and Datasets 2. CitationIE: Leveraging the Citation Graph for Scientific Information Extraction 3. Content-Based Citation Recommendation and 4. unarXive 2022: All arXiv Publications Pre-Processed for NLP, Including Structured Full-Text and Citation Network. However, the justifications provided for these papers highlight that paper no 1. and 3. introduce an approach for citation recommendations during the writing phase of the target manuscripts and not at the proposal writing stage. Also, scientific article 2. leverages contents of a target paper and citation graph to extract scientific information. The outcome of the scientific article 4.is a dataset which can be useful for the proposal, but does not address the task of `topic-based citation retrieval for research proposal'. Thus, we observe that after evaluating the retrieved scientific articles claimed to be invalidating the proposal, the researcher disagrees with the justifications provided for each of the retrieved articles for addressing the motivation behind the proposal, hence validating the novelty of the proposal. This exemplifies the need as well as the effectiveness of the user interaction facility provided by the tool for the workflow. This example demonstrates acceleration of motivation validation stage of the research-life cycle ($\sim$8x for this proposal as mentioned by the researcher),  by eliminating the need for the researcher to manually go through multiple relevant research articles retrieved by generic or academic search engines to ensure that the literature does not have a solution for the specific problem the researcher is trying to address, leading to a time consuming process.

We receive input from another researcher with the proposal titled `\textit{Reference-Free evaluation metric for Retrieval augmented question answering task}' and the abstract `\textit{We observe that questions with long answers on long documents do not have unique reference evidences (relevant paragraphs from the document) and answers. Rather, there is a distribution over reference answers, making expert based evaluation expensive and existing unique reference-based evaluation metrics inadequate. We also do not find any reference-free evaluation metric designed for evaluating retrieval augmented question answering task. Hence, this this work we propose to define this metric.}'. The colleague LLM agent generates the following question to validate the motivation of the proposal: Is the research paper proposes a reference-free evaluation metric designed for evaluating retrieval augmented question answering tasks?. We observe that out of top-50 retrieved scientific articles relevant to the proposal none of the articles provides answer as `yes' to the question,  leading to retrieval of no paper which invalidates the motivation of the proposal. Manual analysis of the top-50 retrieved articles performed by the researcher (as well as other relevant articles manually visited by the researcher) to evaluate this outcome of the workflow, substantiates the results. 

For the next workflow of method synthesis for the above proposal, the mentor LLM agent generates following set of research problems similar to the problem defined in the proposal: 1 Evaluating complex tasks where there is no unique correct answer or reference. 2. Designing evaluation metrics for tasks that involve retrieval and interpretation of large amounts of data. 3. Creating reference-free evaluation metrics for tasks where reference-based metrics are inadequate or impractical. 4. Assessing the quality of answers in tasks where the answers can be long and drawn from extensive documents. The mentor LLM agent also generates the following sub-tasks for the problem defined in the proposal: 1. Defining a new metric that can effectively evaluate retrieval augmented question answering tasks. and 2. Overcoming the inadequacy of existing unique reference-based evaluation metrics for questions with long answers on long documents. With these similar and sub-problems as queries, the colleague LLM agent first retrieves Top-10 similar scientific articles per statement (total 40 articles with some overlap as same article may be retrieved for multiple queries) and then poses the question that `if the article provides a methodology or an approach to solve the above defined problem or sub-task'. The researcher receives total 17 scientific articles which answers `yes' to the question along with  a description of the methodology implemented for each of the above problems or task. Out of these papers, the researcher accepts the following 11 scientific articles, finding them to be more relevant to the problem the researcher is trying to address: 1. AVA: an Automatic eValuation Approach to Question Answering Systems. 2.Evaluation: from precision, recall and F-measure to ROC, informedness, markedness and correlation. 3. Re-visiting Automated Topic Model Evaluation with Large Language Models. 4. SacreROUGE: An Open-Source Library for Using and Developing Summarization Evaluation Metrics. 5. Quantified Reproducibility Assessment of NLP Results. 6. Revisiting the Gold Standard: Grounding Summarization Evaluation with Robust Human Evaluation. 7. A Critical Evaluation of Evaluations for Long-form Question Answering. 8. Think you have Solved Direct-Answer Question Answering? Try ARC-DA, the Direct-Answer AI2 Reasoning Challenge. 9. More Than Reading Comprehension: A Survey on Datasets and Metrics of Textual Question Answering. 10. Hurdles to Progress in Long-form Question Answering. 11. A Critical Evaluation of Evaluations for Long-form Question Answering. 

Based on approaches discussed in these scientific articles to solve the above enlisted similar problems and/or sub-tasks, the mentor LLM agent suggests the following 10 plausible methodologies for the proposal. Theses suggestions with the exact verbatim provided by the agent are as follows: 1. AVA (Automatic Evaluation Approach): This approach uses Transformer-based language models to encode the question, answer, and reference text. By measuring the similarity between the reference and an automatic answer, biased towards the question semantics, AVA can estimate the system accuracy. This approach can be used to evaluate retrieval augmented question answering tasks and overcome the inadequacy of existing unique reference-based evaluation metrics.  2.	Informedness and Markedness Measures: These new probabilistic and information theoretic measures can be used to evaluate complex tasks where there is no unique correct answer or reference. They can also be used to design evaluation metrics for tasks that involve retrieval and interpretation of large amounts of data. 3.	Large Language Models (LLMs): LLMs can be used as automated metrics for evaluating tasks that involve retrieval and interpretation of large amounts of data. They can also be used to assess the quality of answers in tasks where the answers can be long and drawn from extensive documents. 4.	SacreROUGE: This open-source library dedicated to the development of summarization evaluation metrics can be used to develop reference-free evaluation metrics for tasks where reference-based metrics are inadequate or impractical. 5. Quantified Reproducibility Assessment (QRA): This method, based on concepts and definitions from metrology, can be used to develop reference-free evaluation metrics for tasks where reference-based metrics are inadequate or impractical. 6.	RoSE Benchmark: This benchmark can be used to provide a more robust evaluation across different tasks. It can also be used to develop reference-free evaluation metrics for tasks where reference-based metrics are inadequate or impractical. 7.	GENIE6: This human-in-the-loop leaderboard framework can be used to score answers in tasks where the answers can be long and drawn from extensive documents. 8.	Sparse Attention and Contrastive Retriever Learning: This system conditions answer generation on identified documents and uses a dense retriever trained by scaling up a distantly supervised algorithm. It can be used to overcome the inadequacy of existing unique reference-based evaluation metrics for questions with long answers on long documents. 9. Unified Evaluation Benchmark for Long-form Answers: This approach involves conducting a thorough study of evaluation, including both human and automatic evaluation protocols. It can be used to overcome the inadequacy of existing unique reference-based evaluation metrics for questions with long answers on long documents.  10. Training an LFQA Evaluation Metric Directly on Human-Annotated Preference Judgments: This approach involves fine-tuning pre-trained Language Models based on human judgement scores for the task.  This output showcases the quality of method recommendations provided by the tool for the given proposal. Though mentioned at high-level, the researcher agrees that  most of these methods are well-suited as a plausible approach for the given proposal. Though there is a need for further work to finalize the most appropriate plausible method for proposal, the researcher finds this first cut of output provided by the tool to be relevant and the overall process to be  $\sim$10 times more efficient than the regular process followed by the researcher for constructing a plausible set-of approaches for the given problem, by searching through the relevant literature from scratch. 

These examples illustrating the outcomes of the motivation validation and method synthesis phases of the ideation workflow of the tool, demonstrates the efficacy of the tool, in terms of providing relevant outputs at each stage of the workflow. The observations made in terms of time saved by the researchers with the tool usage for the respective tasks demonstrates the power of the tool with regards to time efficiency gains.  


\section{Conclusion}
In this work, we have demonstrated a tool called `Acceleron', developed to accelerate the ideation phase of the research life-cycle. To the best of our knowledge this is the first tool which addresses the tasks involved in the ideation stage. To emulate the ideation process, we use LLM agents with colleague and mentor personas to execute two workflows, viz. motivation validation and method synthesis, which engage researchers in an interactive fashion to develop the research proposal. Our workflow involves novel components to (i) alleviate the hallucinations of LLMs, (ii) ensure relevant outcomes by two-stage aspect based retrieval, where first stage introduces higher recall reducing False Negatives and False Positives are corrected by user interaction, second stage of more precise fine-grained aspect-based retrieval and introduction of unanswerability. The qualitative analysis performed with three proposals from distinct researchers, in the domain of Machine Learning and Natural Language Processing, demonstrates precise outcomes for various stages in the workflow with $\sim$7.5x gains in the time efficiency for various stages of the ideation phase. 

\section{Future Works}
This is an ongoing work. We plan to augment the ideation functionality of the tool for other scientific domains such as life-sciences or material sciences, etc. We plan to emulate the domain specific aspects of the ideation process for these domains, which may differ from the current process specifically designed for Machine Learning and Artificial Intelligence based research projects in Computer Science.  This would allow us to define a meta-process for ideation, which is domain independent and domain specific instances of this meta-process. Currently `Acceleron' supports OpenAI  models. We plan to augment the tool with Open-Source LMs such as Llama-2\cite{Touvron2023Llama2O}, Zypher\cite{tunstall2023zephyr}, etc. The logging functionality of `Acceleron' keeps track of every input provided to the researcher as well as LLM agents and every output from them along with the corresponding timestamps. We are saving these logs for each user interactions for all the sessions. We plan to use these logs with treating user validated inputs as ground truth annotations, to develop a datasets for the ideation process. The logs would be used for developing datasets for tasks such as: (i) retrieval of research papers with similar motivation (ii) proposal re-writing with addressing research-gaps (iii) retrieval of research papers with similar problems and/or (iv) method-synthesis from a set-of relevant papers. The datasets will be used to instruction-tune the above mentioned Open-Source LMs, which can replace the existing LLMs yielding more cost-effective solution. We plan to extend the implementation of current ideation phase to generate a list of experiments to be performed for the problem defined in the proposal and the methodology selected by the researcher. This would lead to generation of a (set-of) results table(s) in a semi-automated fashion, with baseline approaches, planned experiments (ablations) and appropriate metric(s) used for evaluation.

\bibliography{aaai24}

\onecolumn
\section{Appendix}
\subsection{Prompts for different stages of the Workflows}\label{sec:prompts}

\begin{tcolorbox}[title=1. Motivation Extraction Prompt]
System Message:\\
You are a researcher and trying to understand the following proposal written by another researcher:\{proposal\}\\
\\
Human Message:\\
Describe in a bulleted list what is not addressed in the current literature which serves as the Motivation behind solving the above research problem proposed in the Proposal. Answer without a heading line and just the bullet points. Each bullet should mention one gap in the literature as a bullet point and not a sentence.
\end{tcolorbox}
\begin{tcolorbox}[title=2. Motivation Question Generation Prompt]
System Message:\\
You are a researcher and trying to understand the following proposal written by another researcher:\{proposal\}\\
\\
Human Message:\\
Describe in a bulleted list what is not addressed in the current literature which serves as the Motivation behind solving the above research problem proposed in the Proposal. Answer without a heading line and just the bullet points. Each bullet should mention one gap in the literature as a bullet point and not a sentence.\\
\\
AI Message:\\
\{motivation\}\\
\\
Human Message:
Convert each of the above bullets in to a binary question. The question should begin with 'Is the research paper'.
\end{tcolorbox}
\begin{tcolorbox}[title=3. Ask Question for Motivation Validation Prompt]
System Message:\\
You are a researcher. You have been given a context, which are paragraphs from a research paper. You have been given a question. Answer the given Question in 'Yes' OR 'No' OR 'Unanswerable'. Answer solely based on the provided context of the research paper. If the question can not be answered with the facts mentioned in the available context or there is any ambiguity in answering the question answer as 'Unanswerable'. \\
Answer as 'Yes' only when the question can be very clearly answered considering the facts in the research paper provided in the context. Do not repeat the question as the part of the answer. \\
Provide a concise explanation about how the answer to the question is 'Yes' mentioning the paragraphs used in the context to answer it as ‘Yes’. If the answer is 'No' or 'Unanswerable' only output that with NO description or elaboration.\\
\\
Human Message:\\
Question: \{question\} \\
Research Paper Context: \{paper\_chunks\}
\end{tcolorbox}
\begin{tcolorbox}[title=4. Extract Limitation Prompt]
System Message:\\
You are a researcher. You have been given the following proposal: \{proposal\} \\ \\
A different research paper provided in the context already addresses the gap mentioned as the motivation behind the proposal. \\
\{descriptions\} \\
\\
Human Message:\\
Research Paper: \{paper\_chunks\} \\ \\
Identify the limitations or gaps of this research paper which can serve as the new motivation for the proposal. Provide a bulleted list of limitations, where each bullet is concise. Answer WITHOUT a heading line and just the bullet points.
\end{tcolorbox}
\begin{tcolorbox}[title=5. Re-write Research Proposal Prompt]
System Message:\\
You are a researcher and have written a proposal: \{proposal\} \\
\\
Human Message:\\
Re-write the proposal by taking into consideration the mentioned gaps in the current literature as the new motivation behind of the problem defined in the proposal. \\
Answer in a Single detailed paragraph WITHOUT any bullet points or list. \\
Gaps in the current literature: \{limitations\}
\end{tcolorbox}
\begin{tcolorbox}[title=6. Research Problem Extraction Prompt]
System Message:\\
You are a researcher and trying to understand the following proposal written by another researcher:\\
\{proposal\} \\
\\
Human Message:\\
What is the problem solved in the proposal?
\end{tcolorbox}
\begin{tcolorbox}[title=7. Similar Problem Generation Prompt]
System Message:\\
You are a researcher and trying to understand the following proposal written by another researcher:\\
\{proposal\}\\
\\
Human Message:\\
What is the problem solved in the proposal?\\
\\
AI Message:\\
\{problem\_statement\}\\
\\
Human Message:\\
Give me a bulleted list of a more generalised or similar problems to the problem defined in the proposal. Don't give a heading just the answer in a bulleted list.
\end{tcolorbox}

\begin{tcolorbox}[title=8. Sub Problem Generation Prompt]
System Message:\\\
You are a researcher and trying to understand the following proposal written by another researcher: \\
\{proposal\} \\
\\
Human Message:\\
What is the problem solved in the proposal?\\
\\
AI Message:\\
\{problem\_statement\}\\
\\
Human Message:\\
Provide a bulleted list of sub-problems or sub-tasks involved to solve the problem. Don't give a heading just the answer in a bulleted list.
\end{tcolorbox}

\begin{tcolorbox}[title=9. Similar and Sub Problem Question Creation Prompt]
Human Message:\\
\{statement\}\\
For the statement given above generate a question to be posed on a research paper to find out if the paper is proposing an approach or method to perform the task defined by the statement. Start the question with: 'Is the research paper proposing an approach or method to'.
\end{tcolorbox}

\begin{tcolorbox}[title=10. Methodology Extraction Prompt]
System Message:\\
You are a researcher and trying to answer the question posed on a research paper provided as the context.\\
Research Paper: \{paper\_chunks\}\\
\\
Human Message:\\
Answer the given Question in 'Yes' OR 'No' OR 'Unanswerable'. Answer solely based on the provided context of the research paper. If the question can not be answered with the facts mentioned in the available context or there is any ambiguity in answering the question, answer as 'Unanswerable'. Answer as 'Yes' only when the question can be very clearly answered considering the facts in the research paper provided in the context. Do not repeat the question as the part of the answer. If the answer to the question is 'Yes', provide detailed  approach or methodology to perform the task. If the answer is 'No' or 'Unanswerable' only output that with NO description. \\
\\
Question: \{question\}
\end{tcolorbox}

\begin{tcolorbox}[title=11. Method Synthesis Prompt]
System Message:\\
You are a researcher and have been given a proposal and the research problem the proposal is trying to solve. You have been given the approaches in the literature trying to solve, similar problems and sub problems or sub tasks of the problem defined in the proposal. Your task is to synthesize and propose a possible set of methods or approaches to solve the problem defined in the proposal. \\
Proposal: \{proposal\} \\
Research Problem in the Proposal: \{problem\}\\
\\
Human Message:\\
\{method\_context\} \\
\\
Based on the above information suggest the top 3 possible methods or approaches to solve the problem defined in the proposal.
\end{tcolorbox}

\end{document}